\documentclass[letterpaper]{article} 
\usepackage{aaai2027}  
\usepackage[hyphens]{url}  
\usepackage{graphicx} 
\usepackage{natbib}  
\usepackage{caption} 
\usepackage{algorithm}
\usepackage{algorithmic}

\usepackage{newfloat}
\usepackage{booktabs}
\usepackage{listings}
\DeclareCaptionStyle{ruled}{labelfont=normalfont,labelsep=colon,strut=off} 
\floatstyle{ruled}
\newfloat{listing}{tb}{lst}{}
\floatname{listing}{Listing}

\usepackage{booktabs}

\usepackage{amsmath,amssymb}

\usepackage{cleveref} 
\crefname{figure}{Figure}{Figures}
\crefname{table}{Table}{Tables}
\crefname{equation}{Equation}{Equations}
\crefname{section}{Section}{Sections}
\crefname{subsection}{Section}{Sections}
\crefname{algorithm}{Algorithm}{Algorithms}
\crefname{appendix}{Appendix}{Appendices}
\crefname{lemma}{Lemma}{Lemmas}
\crefname{theorem}{Theorem}{Theorems}

\newtheorem{theorem}{Theorem}
\newtheorem{proposition}[theorem]{Proposition}
\newtheorem{lemma}[theorem]{Lemma}
\newtheorem{corollary}[theorem]{Corollary}

\usepackage{amsmath}

\title{Cumulative Marginal Matching for One-Step Diffusion}
\author{
    Paribesh Regmi, Rui Li
}
\affiliations{
    \texttt{\{pr8537, rxlics\}@rit.edu} \\
    Rochester Institute of Technology\\
    Rochester, NY, US
}

\begin{document}

\maketitle

\begin{abstract}
The sampling procedures of diffusion and flow-matching models involve hundreds of denoising steps incurring substantial inference time, which has motivated a growing body of work on training schemes for few-step and one-step generation. Existing approaches primarily seek to align the generation trajectory of few-step and base models, yet their training objectives only involve discrepancies at a single point in the trajectory, encouraging only local \emph{point-level} alignment. While recent methods such as \emph{Shortcut-CSL} introduce cumulative objective that enforce alignment over the entire trajectory, this is suboptimal as they do not  directly enforce alignment between the distributions induced by the two models throughout the generation process. We propose \emph{Cumulative Marginal Matching} (CMM), a principled objective for few-step generation: by viewing flow-matching as a sequence of intermediate marginal distributions connecting the prior and data distributions, CMM directly optimizes the cumulative discrepancy between the base and few-step marginal distributions over the entire sequence. Specifically, we formulate this discrepancy using the Wasserstein-2 distance. By exploiting the pushforward representation of flow-matching, we transform an intractable cumulative Wasserstein objective into a tractable surrogate that can be optimized using samples from the intermediate marginal, which are readily generated through the conditional interpolation process of flow-matching. Furthermore, we show that the existing \emph{Shortcut-CSL} objective arises as a first-order approximation of that of the CMM, establishing a theoretical connection between the two objectives. Experiments on standard benchmarks demonstrate improvement in generation quality over \emph{Shortcut-CSL} and other state-of-the-art one- and few-step baselines under identical training budgets.
\end{abstract}


\section{Introduction}
Modern generative models have largely adopted diffusion \cite{Sohl2015:Deep, Ho2020:DDPM, Song2021:Score-based, Rombach2022:LDM} and flow-matching processes \cite{Lipman2023:Flow, Albergo2023:Stochastic} as their underlying generation paradigm, transforming a simple prior distribution, e.g., Gaussian noise, into a target data distribution. These methods demonstrate remarkable capability in generating structured data such as images, videos and graphs \cite{ho2022:Imagen, gupta2024photorealistic, Eijkelboom2024:VFM, Rombach2022:LDM, vignac2022digress, liu2023generative}, but diffusion-based processes have an inherent efficiency limitation. Both diffusion and flow-matching models require simulating an integral to generate samples, which in practice requires an iterative procedure that gradually transforms Gaussian noise into data samples. As a result, high-quality generation typically requires hundreds of denoising steps. This limitation has motivated a growing body of research on developing methods capable of producing high-quality samples in just one or a few denoising steps \cite{Salimans2022:PD, Yin2024:DMD, Song2023:CM, Frans2025:shortcut, Geng2025:MeanFlow}.

\begin{figure}
\centering
\includegraphics[width=0.40\textwidth]{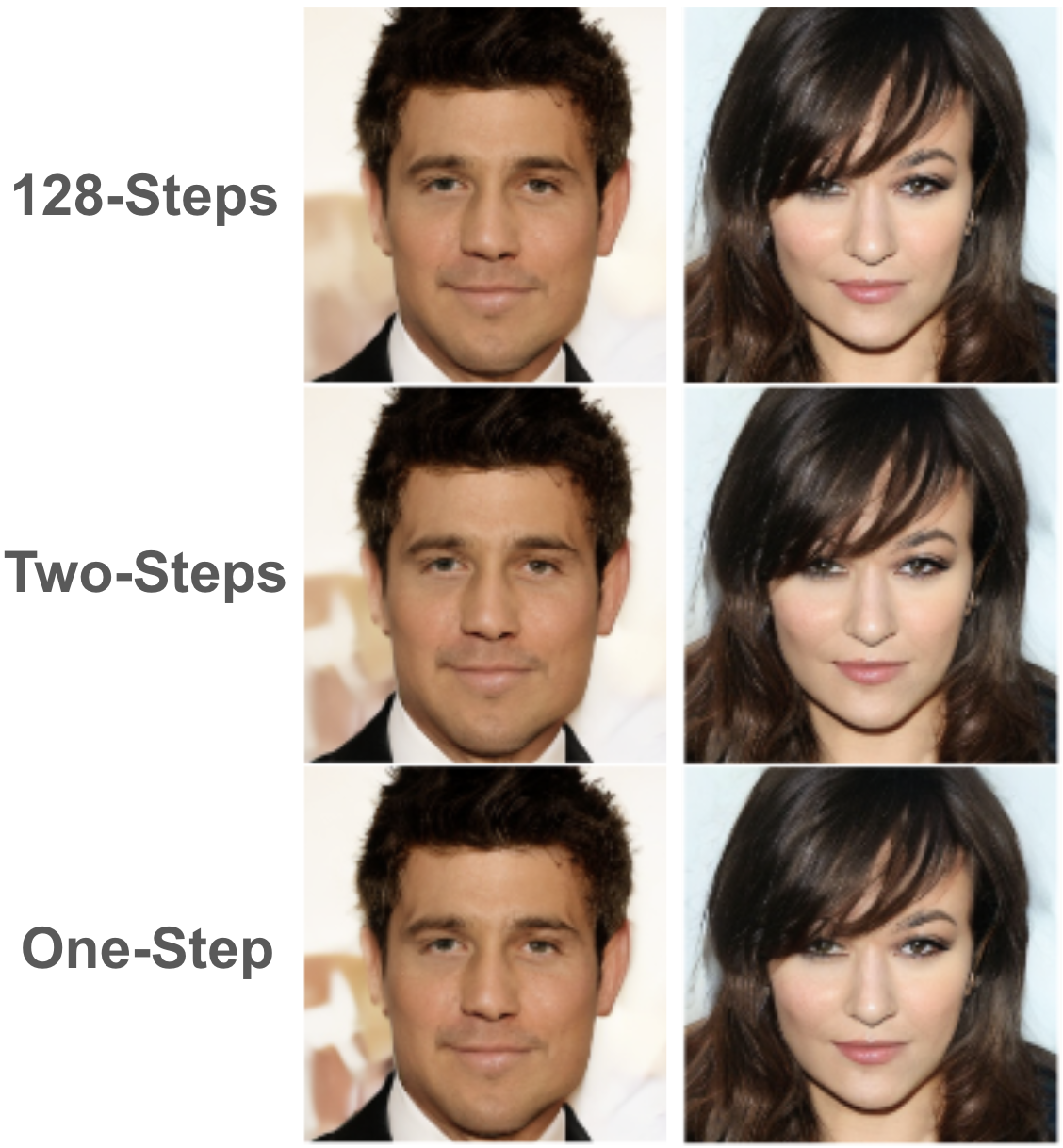}%
\caption{Samples generated from CMM model trained on CelebA-256 with 128, two and one-step sampling. The two- and one-step preserve the sample quality of the base 128-step model.}
\label{fig:samples}
\end{figure}

Methods for one- and few-step generation can be broadly categorized into two classes: two-phase pretrain-and-distill approaches and single-phase approaches that jointly train the base diffusion/flow-matching models and few-step models. In two-phase methods, a high-quality base diffusion or flow-matching model is first trained to learn the flow map from the prior to the data distribution. The knowledge encoded in this transformation is then distilled into a student (few-step) model capable of generating samples directly in one or a few steps \cite{Salimans2022:PD, Yin2024:DMD, Liu2023:Reflow, kim2025:SimpleReflow, Lee2024:ImpReflow}. In contrast, recent single-phase approaches parameterize both the base and few-step generators using a shared backbone network and combine their training by jointly optimizing flow-matching and distillation objectives. Representative examples include Consistency Training \cite{Song2023:CM, Song2024:Improved, Lu2024:SimplifiedCM}, which enforces mapping consistency across different points of the generation process; Inductive Moment Matching \cite{Zhou2025:IMM}, which aligns the base and few-step models through moment-matching objectives; Shortcut Models \cite{Frans2025:shortcut, Nguyen2025:improved, Lin2026:Shortcutting}, which learn multi-step transitions to enable generation with a variable number of denoising steps; and MeanFlow \cite{Geng2025:MeanFlow}, which distills the base model's dynamics by matching the average flow in the generation trajectory. All of these methods formulate distillation as alignment between the teacher and student trajectories, differing primarily in the extent of supervision. Existing methods' distillation objectives are defined through discrepancies evaluated at individual time points, resulting in what we refer to as \emph{point-level objectives}. More recently, Shortcut-CSL \cite{Regmi2025:ST-CSL} extends this formulation by accumulating discrepancies over the entire trajectory, giving rise to a \emph{trajectory-level objective}, encouraging each generation step to account for its impact on future steps. However, we find that such a cumulative objective is suboptimal as it does not precisely target distribution-level discrepancies between the base and few-step models.

From a flow-matching perspective, both the base and few-step generators can be viewed as transporting a common prior distribution through a sequence of intermediate marginal distributions toward the data distribution. This perspective naturally shifts the focus of distillation to matching the distributional evolution itself. In this work, we propose \emph{Cumulative Marginal Matching} (CMM), a principled objective that encourages alignment between the intermediate marginals induced by the base and few-step models throughout the generation process. Specifically, we formulate the objective as the cumulative Wasserstein-2 discrepancy between the corresponding marginals, and show that the pushforward structure of flow matching naturally yields a tractable upper bound that is efficiently optimized using only samples from the intermediate marginal, which are readily generated through the
conditional interpolation process of flow-matching. Furthermore, we establish a theoretical connection between the existing \emph{Shortcut-CSL} objective and CMM, showing that the former can be interpreted as a first-order approximation of the latter. This reveals that, while \emph{Shortcut-CSL} introduces a long-horizon trajectory-level objective, it is not a precise target for matching the distributional evolution of the models.

We make the following three main contributions:
\begin{itemize}
    \item We propose \emph{Cumulative Marginal Matching} (CMM), a principled objective for few-step generation. CMM measures the cumulative discrepancy between the sequence of intermediate marginals induced by the base and few-step models, and we derive a tractable upper bound that enables efficient optimization.
    \item We show that the existing Shortcut-CSL objective arises as the first-order approximation of CMM, providing a theoretical explanation of its relationship to the proposed objective.
    \item We demonstrate the effectiveness of CMM on standard benchmarks. We improve few-step generation quality over state-of-the-art one- and few-step baselines under identical training budgets.
\end{itemize}

\section{Related Works}

Early approaches on accelerating diffusion-based generation primarily follow a two-stage paradigm, where a high-quality teacher model is first trained and subsequently distilled into a student model capable of generating samples in significantly fewer steps. Representative examples include Progressive Distillation \cite{Salimans2022:PD}, which recursively reduces the number of denoising steps through teacher-student distillation, ReFlow \cite{Liu2023:Reflow}, which improves generation efficiency by learning straighter transport trajectories, and Distribution Matching Distillation \cite{Yin2024:DMD}, which directly distills a pretrained diffusion model into a one-step generator with distribution-matching objective. More recent approaches adopt a single-stage training paradigm that jointly optimizes the base model and few-step models within a unified framework. Examples include Consistency Models \cite{Song2023:CM, Song2024:Improved}, which enforce consistency across different points in the generation process; Inductive Moment Matching \cite{Zhou2025:IMM}, which aligns teacher and student dynamics through moment-matching objectives; MeanFlow \cite{Geng2025:MeanFlow}, which distills generation dynamics by matching the average teacher flow; and Shortcut Models \cite{Frans2025:shortcut}, which train a single network to predict larger-step transitions and enable flexible one- and few-step generation, and flow-map formulation for few-step generation \cite{Boffi2025:FlowMap, Boffi2024:FlowMM}, which generalizes MeanFlow and Consistency Models by flow-map formulation. These methods largely rely on objectives that enforce local consistencies, i.e. consistency along a single point in the trajectory, which are independent of one another. Such objectives encourage few-step models to take steps that are only locally aligned with the base model, ignoring how current step affects the subsequent steps along the trajectory.

Recent work has proposed training few-step models with a trajectory level objective that encourages few-step models to account for the downstream effects as well. \emph{Shortcut-CSL} \cite{Regmi2025:ST-CSL} put forward a perspective of viewing few-step models through the lens of optimal control and propose a novel objective that cumulatively penalizes misalignment between the base and the few-step models along the entire trajectory. This leads to improved one- and few-step generation performance compared to point-level distillation objectives. However, we find that the \emph{Shortcut-CSL} objective is suboptimal as it does not directly enforce agreement between the distributions induced by the base and few-step models over time. To this end, our work considers the evolution of the intermediate marginal distributions themselves and proposes a cumulative marginal matching objective that encourages distributional alignment throughout the entire generation process.

\section{Background}

\subsection{Flow Matching}

Flow-matching \citep{Lipman2023:Flow, Albergo2023:Stochastic} constructs a map from an easy-to-sample reference distribution (typically an Normal Gaussian) to a complex target distribution supported on high-dimensional data. Specifically, it learns a velocity field $v_\theta$ that transports samples from the reference distribution to the data distribution.

The velocity field is trained by minimizing
\begin{align}
\mathcal{L}_{FM}(\theta) = \mathbb{E}_{x_0,x_1} \left[ \|v_\theta(x_t,t)-v_t\|^2 \right]
\end{align}

Under the linear interpolation regime \citet{Lipman2023:Flow}, the noise sample $x_0 \sim \mathcal{N}(0,I)$ and a data sample $x_1 \sim D$ are connected by
\begin{align}
\label{eq:fm_obj}
x_t=(1-t)x_0+t x_1, \qquad v_t=x_1-x_0
\end{align}

After training, samples are generated by integrating the learned ODE
\begin{align}
dx_t=v_\theta(x_t,t)\,dt,\qquad
x_0\sim\mathcal{N}(0,I).
\end{align}

Although each training trajectory is a straight line, the model learns a single global velocity field that must simultaneously represent the transport induced by all training pairs. Since neighboring points generally belong to different straight-line trajectories with different directions, the learned field varies smoothly across space, causing its integral curves to be generally curved.

\subsection{Few-Step Model and Trajectory-Level Objective}

\begin{figure*}[]
\centering
\includegraphics[width=0.90\textwidth]{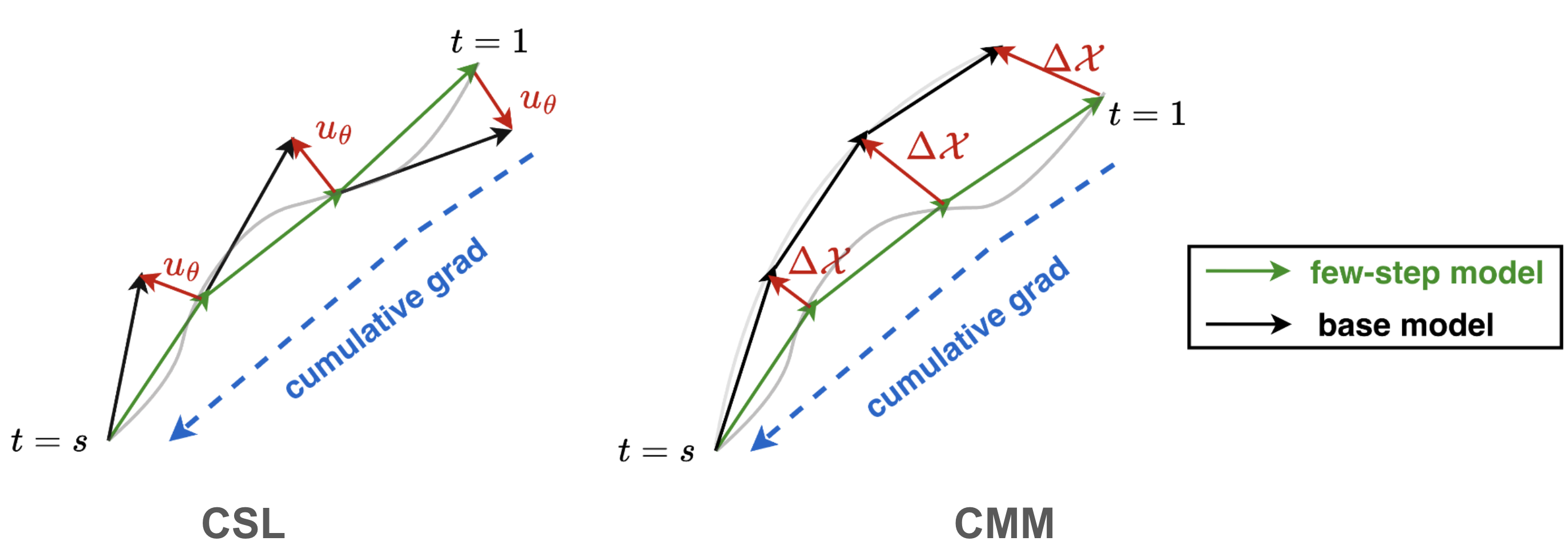}%
\caption{Illustration of the CSL (left) and CMM (right) losses (the illustration of CSL is recreated based on the original \emph{Shortcut-CSL} paper). In the CSL loss, few-step model is unfolded in time and its  misalignment ($u_\theta$) with the model velocity field is penalized at each point. In CMM, we let the base and few-step model unfold together on their own trajectories, and penalize the difference between their trajectories ($\Delta X$) at each point.}
\label{fig:cmm_loss}
\end{figure*}

The curved trajectories learned by flow matching require many integration steps for accurate sampling. Shortcut models \citep{Frans2025:shortcut} address this by learning a step-size-conditioned velocity field that directly advances the trajectory over an interval of length $d$:
\begin{align}
\label{eq:x_t_d}
x_{t+d}=x_t+\mathcal{S}_\theta(x_t,t,d)\cdot d.
\end{align}
When $d=0$, the model reduces to the standard flow-matching velocity field.

Since no supervision exists for large-step predictions, shortcut models exploit the self-consistency of trajectories: one step of size $2d$ should produce the same state as two successive steps of size $d$. This yields the self-consistency target
\begin{align}
\mathcal{S}_{\mathrm{target}}
=
\frac{1}{2}\mathcal{S}_\theta(x_t,t,d)
+
\frac{1}{2}\mathcal{S}_\theta(x_{t+d},t+d,d),
\end{align}
which is optimized jointly with the standard flow-matching objective.
\begin{align}
\mathcal{L}_{\text{shortcut}}(\theta) = \mathbb{E}_{x_0, x_1, t, d} 
&\Big[
\underbrace{\| \mathcal{S}_\theta(x_t, t, 0) - (x_1 - x_0) \|^2}_{\text{Flow-Matching}}\quad + \nonumber \\ &\underbrace{\| \mathcal{S}_\theta(x_t, t, 2d) - \mathcal{S}_{\text{target}} \|^2}_{\text{Self-Consistency}}
\Big]
\end{align}
\emph{Shortcut-CSL} \cite{Regmi2025:ST-CSL} view few-step generation as a controlled dynamical system,
\begin{align}
dX_t^u=\big[b(X_t^u,t)+u_\theta(X_t^u,t)\big]dt,
\end{align}
where
$
u_\theta(X_t^u,t)=\mathcal{S}_\theta(X_t^u,t)-b(X_t^u,t)
$
denotes the deviation of the few-step model from the base flow-matching velocity field. Under this interpretation, they propose the cumulative self-consistency (CSL) objective,
\begin{align}
\label{eq:csl_loss}
J_{CSL}(u_\theta;X_s^u,s)
=
\int_s^1
\|u_\theta(X_t^u,t)\|^2dt,
\end{align}
which penalizes the accumulated trajectory-wise deviation from the base model. They further show that the original shortcut self-consistency loss is an instantaneous approximation of this cumulative objective at $t=s$,
\begin{align}
\label{eq:sl_loss}
&J_{SL}(u_\theta;X_s^u,s) \nonumber \\
= &\|u_\theta(X_s^u,s)\|^2 \nonumber \\
= &\|\mathcal{S}_\theta(X_s^u,s,2d)- \mathcal{S}_{\mathrm{target}}\|^2
\end{align}

\section{Cumulative Marginal Matching}
Flow Matching defines a family of time-indexed marginal distributions $\{p_t(x)\}_{t\in[0,1]}$ that interpolate between a simple prior distribution $p_0$ and the data distribution $p_1$. These marginals are constructed by specifying a conditional path distribution $p_t(x \mid x_1)$, which describes how a data point $x_1$ is transported in time.
\begin{align}
\label{eq:fm_marginal}
    p_t(x) = \int p_t(x \mid x_1)\, q(x_1)\, dx_1
\end{align}
The choice of the conditional distribution $p_t(x \mid x_1)$ defines the transport path between the prior and data distributions. The aim is to learn a deterministic velocity field $v_\theta(x,t)$ whose associated flow reshapes the prior distribution to the intermediate marginal distributions $\{p_t(x)\}$ over time, and finally to the data distribution $p_1$. The time-dependent flow is given by:
\begin{align}
    \psi_t(x_0) = x_0 + \int_0^t v_\theta(x_s,s)\, ds ,
\end{align}
With this flow map, the marginal at time $t$ can be considered as the pushforward of the prior under $\psi_t$, i.e. $p_t = [\psi_t]_* p_0$ \cite{Lipman2023:Flow}, which satisfies the relation:
\begin{align}
    dp_t(x_t) = dp_0\left(\psi_t^{-1}(x_t)\right)
\end{align}

Thus, both the base model and the shortcut model can be viewed as transporting a simple base distribution to the data distribution over time. Therefore, when distilling a pretrained model, the natural objective is to preserve the evolution of marginal distributions. To this end, we propose a novel training objective for shortcut models that penalizes discrepancies between the marginals of the base and shortcut models at each time step. Specifically, we consider a cumulative Wasserstein-2 distance between the corresponding marginals over time,
\begin{align}
\label{eq:begin_def}
    \mathcal{L}_W = \int_{s}^1 \, \mathcal{W}_2\left(p_t^u, p_t\right)\, dt
\end{align}

where $p_t$ and $p_t^u$ are the marginal distributions of the base model and the shortcut models respectively. Directly optimizing $\mathcal{L}_W$ is not tractable since the marginal distributions are not available in closed form. However, as we will show, the pushforward representation naturally induces a tractable upper bound.

Let us define the flow maps induced by the base and few-step models from time $s$ to any time $t$ ($s\leq t \leq 1)$ as:
\begin{align}
    X_t=\psi_{t|s}(X_s),\ &\ \
    X_t^u=\psi_{t|s}^{u}(X_s) \\
    \text{and}, X_s\sim p_s
\end{align}
where $\psi_{t|s}$ and $\psi_{t|s}^{u}$ denote the flow map from time $s$ to $t$. 
Let us define the deviation between the base and few-step trajectories by:
\begin{align}
    \Delta X_t = X_t^u-X_t
\end{align}
The flow map satisfies the composition property $\psi_t=\psi_{t|s} \circ \psi_s$. Thus,
\begin{align}
\label{eq:pushforward}
    p_t=[\psi_{t|s}]_*p_s
\end{align}
Since both marginal distributions originate from the same prior distribution, the mapping
\begin{align}
    T = \psi_{t|s}^u \circ \psi_{t|s}^{-1}
\end{align}
transports $p_t$ onto $p_t^u$. Therefore, by Monge's formulation of optimal transport \cite{villani2009:optimal,peyre2019:computational},
\begin{align}
    \mathcal{W}_2^2(p_t^u,p_t) \le \int_{\mathcal X} \|T(X)-X\|^2 \,dp_t(X)
\end{align}

where equality holds if $T(x)$ represents an optimal map. Substituting the above transport map gives
\begin{align}
\label{eq:same_point}
    T(X_t)-X_t = &\psi_{t|s}^u(X_s)-\psi_{t|s}(X_s) \nonumber \\
    = &\Delta X_t,
\end{align}
which yields
\begin{align}
\label{eq:w2_upper}
    \mathcal{W}_2^2(p_t^u,p_t)
    \le
    \int_{\mathcal X}
    \|\Delta X_t\|^2
    \,dp_t
\end{align}

Equation~\eqref{eq:w2_upper} remains intractable because although samples from the intermediate marginal $p_t$ can be generated through the conditional interpolation process in flow-matching (Equation~\eqref{eq:fm_obj}), such samples alone are insufficient to evaluate the transport displacement $\Delta X_t$. Specifically, $\Delta X_t$ is defined as the deviation between the coupled trajectories of the base and few-step models initialized from the same starting point $X_s$(Equation~\eqref{eq:same_point}). Therefore, evaluating this quantity requires access to the corresponding flow trajectories rather than isolated samples from the intermediate marginal.

A key observation is that flow-matching represents every intermediate marginal as the pushforward of the prior distribution under the flow map, as defined in Equation (\ref{eq:pushforward}). Consequently, expectations over the intermediate marginal can be rewritten entirely as expectations over the fixed prior distribution. The following elementary property of pushforward measures \cite{bogachev2007measure} allows us to rewrite
expectations with respect to the intermediate marginal distributions.

\begin{proposition}
\label{prop:pushforward}
Let $\mu=[\psi]_*\nu$, where $\psi:\mathcal X\rightarrow\mathcal X$ is measurable.
Then, for any measurable function $f:\mathcal X\rightarrow\mathbb R$,
\begin{align}
\label{eq:pushforward_expectation}
    \int_{\mathcal X} f(x)\,d\mu(x) &= \int_{\mathcal X} f(\psi(x)) \,d\nu(x)
\end{align}
\end{proposition}

Applying Proposition~\ref{prop:pushforward} to \cref{eq:w2_upper} with $f(X)=\|T(X)-X\|^2$ gives:
\begin{align}
\mathcal W_2^2(p_t^u,p_t) &\le \int_{\mathcal X} \|T(X)-X\|^2 \,dp_t(X) \nonumber \\
&= \int_{\mathcal X} \left[ \|T(\psi_{t|s}(X_s))-\psi_{t|s}(X_s)\|^2 \right] \,dp_s(X_s) 
\end{align}
Since $T(\psi_{t|s}(X_s)) = \psi_{t|s}^u(X_s)$, we obtain:
\begin{align}
T&(\psi_{t|s}(X_s)) - \psi_{t|s}(X_s) \nonumber \\
=\ &\psi_{t|s}^u(X_s) -  \psi_{t|s}(X_s) \nonumber \\ 
=\ &\Delta X_t
\end{align}
We see that this change of variables only reparameterizes the expectation; the transport displacement itself remains unchanged and is precisely the trajectory difference $\Delta X_t$ between the base and few-step flows. Therefore,
\begin{align}
\label{eq:above}
\mathcal W_2^2(p_t^u,p_t) \leq \mathbb E_{X_s\sim p_s} \left[ \|\Delta X_t\|^2 \right]
\end{align}

This reformulation is particularly appealing since it removes the dependence on the intermediate marginal $p_t$ . Instead, the Wasserstein discrepancy is expressed through the deviation between the coupled trajectories of the base and few-step models initialized from the same starting state $X_s\sim p_s$.

Thus, the upper bound to Equation (\ref{eq:begin_def}) is expressed as:
\begin{align}
    \mathcal{L}_{\mathrm W} &\leq \int_s^1 \mathbb E_{X_s\sim p_s}
    \left[ \|\Delta X_t\|^2 \right]dt \nonumber \\
    &= \mathbb E_{X_s\sim p_s} \left[ \int_s^1 \|\Delta X_t\|^2dt \right]
\end{align}

With this, we define the proposed CMM objective:
\begin{align}
\boxed{
    \mathcal L_{CMM} = \mathbb E_{x_s\sim p_s} \left[ \int_s^1 \|\Delta X_t\|^2 dt \right]
}
\end{align}

Although $\mathcal{L}_{CMM}$ is derived through the pushforward representation of the flow, it only requires samples from the intermediate marginal $p_s$ during optimization. In flow-matching, such samples can be efficiently obtained through the conditional interpolation process (as in Equation~\eqref{eq:fm_obj}) without explicitly evaluating the marginal density. Also, starting from the same sampled state $X_s$, the base and few-step models generate coupled trajectories, which allows direct computation of the trajectory deviation $\Delta X_t$. An illustration of the proposed CMM objective is provided in Figure \ref{fig:cmm_loss}, along with side by side comparison with CSL.

\subsection{Relation to CSL}

The following lemma provides an alternative characterization of the CMM objective by decomposing the trajectory discrepancy:

\begin{lemma}
\label{lemma:cmm_decomp}
The CMM objective can be equivalently written as:
\begin{align}
\label{eq:cmm_decomp}
\mathcal{L}_{CMM} = &\mathbb{E} \left[ \int_s^1 (1-t) \Delta X_t^{\top} u_\theta(X_t^u,t) dt \right] + \nonumber \\
&\mathbb{E} \left[ \int_s^1 (1-t) \Delta X_t^{\top} 
\Delta b_t\ dt \right]
\end{align}
where
\begin{align}
\Delta b_t = b(X_t^u,t)-b(X_t,t) \nonumber
\end{align}
\end{lemma}




Lemma~\ref{lemma:cmm_decomp} reveals that the CMM objective accounts for two
sources of discrepancy between the base and few-step trajectories. The first
term measures the discrepancy between base and few-step model caused by the deviation $u_\theta(X_t^u,t)$. The second term
captures the change in the base dynamics caused by evaluating the flow of the base and few-step models at different states. This term is absent in CSL, where the difference between base and few-step model's prediction is evaluated at the same state $X^u_t$ (Figure \ref{fig:cmm_loss}) and therefore does not explicitly account for the evolution of the base flow discrepancy. We formalize this relationship in the following theorem.

\begin{theorem}
\label{thm:csl_local}
Consider an interval $[s,s+d]$ with step size $d$. Define the restricted CMM objective as
\begin{align}
\mathcal{L}_{CMM}[s,s+d] = \mathbb{E}_{X_s\sim p_s} \left[ \int_s^{s+d} \|\Delta X_t\|^2 dt \right]
\end{align}
where $\Delta X_t=X_t^u-X_t$ denotes the deviation between the few-step and base trajectories initialized from the same state $X_s$. Under standard regularity assumptions, the CMM objective admits the following expansion:
\begin{align}
\mathcal{L}_{CMM}[s,s+d] = \frac{d^3} {3} \mathbb{E}_{X_s\sim p_s} \left[ \|u_\theta(X_s,s)\|^2 \right] + \mathcal{O}(d^4)
\end{align}
\end{theorem}

\begin{corollary}
\label{cor:csl_global}
Consider the full CMM objective over the interval $[s,1]$ and assume the generation trajectory is partitioned into intervals of size $d$.
Then,

\begin{align}
\mathcal{L}_{CMM} = \frac{d^2}{3} J_{CSL}(t) + \mathcal{O}(d^3)
\end{align}
\end{corollary}
This observation suggests that CSL arises as an approximation of CMM by only considering the first-order terms and neglecting the higher-order effects arising from the base flow discrepancy. CSL becomes an increasingly accurate approximation of CMM as the step size $d$ decreases, while the accumulated approximation error becomes significant in the few-step regime with large $d$.

\subsection{Estimation of $\mathcal{L}_{CMM}$}

The integral in $\mathcal{L}_{CMM}$ is estimated in discrete time by cumulating the magnitude of the path difference along the entire trajectory. Specifically,
\begin{align}
\mathcal{L}_{CMM}(x_{nd}, nd) = \sum_{k=n}^{R'} \left\| \Delta X_{dk} \right\|^2, \quad R'=\frac{1}{d}
\end{align}
Calculating the full summation requires trajectory simulation. Therefore, following the approximation in \cite{Regmi2025:ST-CSL}, we approximate the objective using only a small number of future steps. he number of terms in the summation is given by $R=(R'+1)-n$. We use $R=2$, which provides a favorable trade-off between computational cost and approximation quality.

The truncated objective is optimized by unrolling the shortcut trajectory over the next $R$ steps and accumulating the corresponding path discrepancies. The second discrepancy is evaluated on a state generated by the first prediction, and since this state itself depends on the first update ($x_{t+d} = x_t + s(x_t, t, d) \cdot d$), the second discrepancy becomes an implicit function of the first prediction through the unrolled computation graph. Consequently, automatic differentiation propagates gradients from the second discrepancy through the first shortcut update, called as the \emph{cumulative gradient}, encouraging the current prediction to reduce not only its immediate trajectory deviation but also the deviation accumulated at subsequent states. More generally, increasing $R$ extends this dependency over a longer horizon, allowing each shortcut prediction to account for its influence on future trajectory discrepancies. A visualization of the cumulative gradient is in Figure \ref{fig:cmm_loss}.

\section{Empirical Analysis}
In this section, we present an empirical study of the performance, scalability, and efficiency of the proposed CMM, comparing it against baseline few-step generation methods. We train our method on CelebA-256~\cite{Liu2015:CelebA} and ImageNet-256~\cite{Deng2009:Imagenet} benchmarks. In the following section, we evaluate the sample quality of both the baselines and our method on these benchmarks. We further examine the scalability of our approach when applied to larger backbone networks. Next, we report the training time of our method and compare it against the baselines. All models are trained in the latent space of a stable VAE~\cite{Rombach2022:LDM}, and generative performance is measured using the FID-50K metric. The backbone network parameterizing the velocity field $v_\theta$ is a medium scale diffusion transformer DiT-B-2 \cite{Peebles2023:DiT} unless stated otherwise. Model training for all experiments are done on Nvidia A100 GPU. Other experimental details and hyperparameters are provided in the Appendix.

\subsection{Performance Evaluation}
\label{sec:performance}

\textbf{Choice of baselines:} To rigorously assess the impact of the proposed CMM objective on generative performance, we ideltify \emph{Shortcut-CSL} (ST-CSL) ~\cite{Regmi2025:ST-CSL} as the closest baseline, as a direct comparison with it isolates and evaluates the effectiveness of the CMM objective itself. To further situate our method within the broader landscape of few-step generative modeling, we additionally compare against the state-of-the-art baselines considered by ST-CSL paper, including the widely adopted MeanFlow~\cite{Geng2025:MeanFlow} method. The other baselines we use are Reflow \cite{Liu2023:Reflow}, Progressive Distillation \cite{Salimans2022:PD} and Consistency Training \cite{Song2023:CM}.

    Table \ref{tab:comparison_all} report the FID scores of baselines and our method under two generation settings: (i) unconditioned generation on CelebA-256 and (ii) class-conditioned generation on ImageNet-256. The results of the baselines are adopted from ST-CSL paper. For \emph{Shortcut}, ST-CSL, and our (CMM) methods, we report the performance for two different $B:K$ ratios (the proportion of number of data points used for flow-matching and distillation part of the objectives, please see appendix for more details). We see that CMM outperforms the baselines on both datasets and across all generation steps except ST-CSL, compared to which our method does better only on four-step generation, and not for two- and one-step generations. However, when we train the  with the combination of CMM and CSL objectives i.e., train all few-step models with CMM, except for two- and one-step models which are trained with CSL, we observe that this combination achieves the best performance at all generation steps. We explain this behavior of CMM and the motivation to combine the two losses with the following theorem, which is related to the Lipschitz properties of the flow map induced by the velocity field $v_\theta$ \cite{stephanovitch2026:lipschitz}.

\begin{table*}[h!]
\centering
\begin{tabular}{lccc|ccc}
\toprule
& \multicolumn{3}{c|}{\textbf{CelebA-256}} & \multicolumn{3}{c}{\textbf{ImageNet-256}} \\
\cmidrule(lr){2-4} \cmidrule(lr){5-7}
\textbf{Method} & \textbf{Four-Step} & \textbf{Two-Step} & \textbf{One-Step} 
& \textbf{Four-Step} & \textbf{Two-Step} & \textbf{One-Step} \\
\toprule
\textbf{Reflow} 
& 13.77{\scriptsize$\pm$0.05} & 14.48{\scriptsize$\pm$0.05} & 16.07{\scriptsize$\pm$0.02}
& 40.86 & 43.68 & 50.40 \\
\textbf{Progressive Distillation} 
& 14.49{\scriptsize$\pm$0.07} & 16.73{\scriptsize$\pm$0.02} & 20.40{\scriptsize$\pm$0.15}
& - & - & - \\
\textbf{Flow-Matching} 
& 62.8{\scriptsize$\pm$0.04} & 112.1{\scriptsize$\pm$0.07} & 321.2{\scriptsize$\pm$0.04}
& 104.96 & 210.39 & 325.78 \\
\textbf{Consistency Training} 
& 19.0 & - & 33.2
& - & - & - \\
\textbf{MeanFlow}
& - & - & -
& 34.09 & 36.61 & 45.12 \\
\midrule
\textbf{Shortcut (4:1)}
& 9.36{\scriptsize$\pm$0.05} & 12.56{\scriptsize$\pm$0.02} & 20.46{\scriptsize$\pm$0.02}
& 24.70 & 35.73 & 64.12 \\
\textbf{ST-CSL (4:1)}
& 8.98{\scriptsize$\pm$0.02} & 10.96{\scriptsize$\pm$0.02} & 18.37{\scriptsize$\pm$0.02}
& 16.98 & 21.77 & 45.84 \\
\textbf{CMM (4:1)}
& 8.30{\scriptsize$\pm$0.03} & 11.26{\scriptsize$\pm$0.03} & 19.48{\scriptsize$\pm$0.05}
& 14.96{\scriptsize$\pm$0.06} & 22.68 {\scriptsize$\pm$0.07} & 48.71{\scriptsize$\pm$0.10} \\
\textbf{CMM+CSL (4:1)}
& \textbf{8.28{\scriptsize$\pm$0.02}} & \textbf{10.70{\scriptsize$\pm$0.06}} & \textbf{17.70{\scriptsize$\pm$0.05}}
& \textbf{14.55{\scriptsize$\pm$0.09}} & \textbf{18.74{\scriptsize$\pm$0.11}} & \textbf{41.83{\scriptsize$\pm$0.12}} \\
\midrule
\textbf{Shortcut (1:1)}
& 9.00{\scriptsize$\pm$0.02} & 11.14{\scriptsize$\pm$0.02} & 16.51{\scriptsize$\pm$0.02}
& 24.17 & 32.22 & 51.78 \\
\textbf{ST-CSL (1:1)}
& 8.54{\scriptsize$\pm$0.02} & 9.91{\scriptsize$\pm$0.02} & 15.51{\scriptsize$\pm$0.02}
& 15.71 & 17.35 & 31.66 \\
\textbf{CMM (1:1)}
& 8.15{\scriptsize$\pm$0.05} & 10.41{\scriptsize$\pm$0.06} & 16.91{\scriptsize$\pm$0.04}
& \textbf{13.85{\scriptsize$\pm$0.08}} & 19.25{\scriptsize$\pm$0.11} & 37.72{\scriptsize$\pm$0.09} \\
\textbf{CMM+CSL (1:1)}
& \textbf{8.08{\scriptsize$\pm$0.03}} & \textbf{9.18{\scriptsize$\pm$0.02}} & \textbf{14.57{\scriptsize$\pm$0.03}}
& \textbf{13.77 {\scriptsize$\pm$0.10}} & \textbf{14.28{\scriptsize$\pm$0.06}} & \textbf{27.19{\scriptsize$\pm$0.13}} \\
\bottomrule
\end{tabular}
\caption{Comparison of the generation quality of the baselines and our method for unconditional generation on CelebA-256, and class-conditional generation on Imagenet-256 in terms of the FID-50K $\downarrow$ scores. The ratios after the methods represent the proportion of the number of data points used for the flow-matching and distillation parts of the objectives.}
\label{tab:comparison_all}
\end{table*}

\begin{theorem} [Lipschitz Constant of the Flow Map]
\label{th:lipchitz}
Let $\psi_t$ denote the flow generated by the velocity field $ v_\theta(x_t,t)$, and suppose that $v_\theta(\cdot,t)$ is $L$-Lipschitz for all $t$, i.e.,
\begin{align}
\|v_\theta(x,t)-v_\theta(y,t)\|
\le L\|x-y\|,
\qquad \forall x,y,t \nonumber
\end{align}
Then the flow map $\psi_t$ is $e^{Lt}$-Lipschitz:
\begin{align}
\|\psi_t(x)-\Phi_t(y)\|
\le e^{Lt}\|x-y\|,
\qquad \forall x,y \nonumber
\end{align}
In particular, for $t\in[0,1]$,
\begin{align}
\operatorname{Lip}(\psi_{0\to1})
\le e^L
\end{align}
\end{theorem}

The proof of the theorem follows straightforwardly from \textit{Gronwall's inequality} and is provided in the Appendix.

\begin{corollary}[Lipschitz Constant of Finite-Step Generation]
\label{cor:lipchitz}
Let  $\psi^{(n)}$ denote the $n$-step Euler approximation of the flow map
over the interval $[0,1]$:
\begin{align}
x_{k+1}
=
x_k + \frac{1}{n}v_\theta(x_k,t_k),
\qquad
t_k=\frac{k}{n} \nonumber
\end{align}
Then
\begin{align}
\operatorname{Lip}(\psi^{(n)})
\le
\left(1+\frac{L}{n}\right)^n \ \ \& \ \ \left(1+\frac{L}{n}\right)^n \le e^L \nonumber
\end{align}

with equality attained in the limit $n\to\infty$. Consequently,
\begin{align}
\operatorname{Lip}(\psi^{(n)})
\le
\operatorname{Lip}(\psi_{0\to1}),
\end{align}
and the Lipschitz constant of the generator decreases as the number of
generation steps is reduced.
\end{corollary}

As the number of generation steps decreases, the Lipschitz upper bound of the induced transport map decreases from $e^L$ to $\left(1+\frac{L}{n}\right)^n$. This results in a progressively more constrained class of transport maps, which may reduce the model's capacity to approximate complex distributional transformations. Importantly, Theorem \ref{th:lipchitz} and Corollary \ref{cor:lipchitz} suggest that the Lipschitz upper bound of the transport map grows monotonically with the number of steps, implying that a four-step model can represent a richer class of transformations than a two-step model.

From this perspective, the CMM objective imposes a substantially stronger constraint compared the CSL objective, as it explicitly requires the few-step model to match the trajectory of its teacher, while the CSL objective, being its biased estimator, is a more relaxed constraint. For fewer generation steps, such as two-step generators, this constraint may exceed the representational capacity of the model, resulting in optimization difficulties and unstable performance. In contrast, four-step models possess a larger Lipschitz budget and are therefore better able to satisfy the CMM constraint, leading to improved performance.

This observation motivates our hybrid training strategy. We apply CMM only to models with sufficient capacity (up to four-step generation), where the stronger supervision is beneficial, while training more aggressively distilled models (two-steps) using CSL alone. In this regime, the two-step model is no longer forced to reproduce behaviors that may lie beyond its representational capacity. Since CMM improves the quality of the four-step model, the two-step model ultimately learns from a stronger teacher than would be available under CSL-only training. This combination yields better performance than either applying CMM uniformly across all shortcut levels or training all models exclusively with CSL.

\subsection{Training Efficiency}
\label{sec:efficiency}
In this section we compare the training times of our method with that of the \emph{Shortcut}, ST-CSL, and Flow-Matching models when trained on the CelebA-256 dataset. All models are trained on Nvidia A100 GPU with the same batch sizes, on diffusion transformers of small (DiT-S-2, 33M) and medium (DiT-B-2, 130M) scales. and the reported wall-clock time for 100 epochs of training is provided in Table \ref{tab:efficiency}.

We see that compared to the \emph{Shortcut} and ST-CSL methods, CMM consumes slightly more training time. This is because while calculating the CMM objective, both the base model and few-step models are rolled out independently, whereas for CSL, only the few-step model is rolled out, as illustrated in Figure \ref{fig:cmm_loss}.


\begin{table}[t]
\centering
\begin{tabular}{lcc}
\multicolumn{3}{c}{\textbf{Training Time} (in hours)} \\
\toprule
\textbf{Method} & \textbf{DiT-S-2} & \textbf{DiT-B-2} \\
\midrule
\textbf{FM}       & --  & 6.4 \\
\textbf{Shortcut} & 5.6 & 7.4 \\
\textbf{ST-CSL}   & 6.2 & 7.8 \\
\textbf{CMM}      & 6.6 & 8.6 \\
\bottomrule
\end{tabular}
\caption{Comparison of the wall-clock training time of Flow Matching (FM), \emph{Shortcut}, \emph{Shortcut-CSL} (ST-CSL), and CMM over 100 training epochs on the CelebA-256 dataset using Diffusion Transformers of different sizes.}
\label{tab:efficiency}
\end{table}

\section{Conclusion}
In this work, we propose a cumulative objective that optimizes the discrepancy between the sequence of marginal distributions induced by the base and few-step models. We further establish theoretical connections between our framework and existing approaches, showing that existing methods can be viewed as approximations of our objective. Experiments on standard generative modeling benchmarks demonstrate that our method consistently improves generation quality, as measured by FID scores. An interesting direction for future work is to empirically investigate the implications of Theorems 0.5 and 0.6, which suggest a fundamental limitation on few-step generation and may explain the performance gap between few-step models and their continuous-time counterparts.

\bibliography{aaai2027.bib}


\onecolumn
\begin{center}
{\LARGE Appendix for Cumulative Marginal Matching for One-Step Diffusion}
\end{center}

\section{Experimental Details}
For experiments in Tables 1 \& 2 in the main paper, we follow the basic hyperparameter setup from the \emph{Shortcut-CSL} paper, which are detailed in Table \ref{tab:hyperparams} below. Due to computational limitations, we follow the setting of \emph{Shortcut-CSL} and sample total of $\sim$40K image samples from 30 classes for class-conditional training on ImageNet-256. $B$ is the number of data points in a batch used for flow-matching loss, $K$ denotes the number of data points used to compute the CMM loss ($\mathcal{L}_{CMM}$), and $R$ is the number of terms used to approximate $\mathcal{L}_{CMM}$. For fair comparison with \emph{Shortcut} and \emph{Shortcut-CSL}, we keep the values of $B$ and $K$ the same in these baselines and our method. For our method and \emph{Shortcut-CSL}, we have used a value of $R=2$. For Meanflow baseline, we report its best performance from \emph{Shortcut-CSL} paper, corresponding to the optimal ratio of flow-matching to bootstrap targets. The evaluation metric is FID-50k, which is essentially the Wasserstein-2 distance between the features of the True and generated images, obtained from the Inception-v2 network. It is a popular metric and widely used to evaluate generative models' quality. The mean and standard deviation of the FID scores from three independent runs are reported.

The number of denoising steps in the base flow-matching model is set to 128. The \texttt{3*256*256} images of the CelebA and ImageNet are compressed to the latent space of VAE \texttt{sd-vae-ft-ema} of size \texttt{3*256*256}. The flow-matching model is trained in this latent space, and during inference, the generate latent codes are fed to the VAE decoder to obtain images.

\begin{table}[htb]
\centering
\begin{tabular}{lcc}
\hline
\textbf{Hyperparam} & CelebA-256 & Imagenet-256 \\
\hline
Dataset Size & $\sim$30k & $\sim$40k \\
\#Epochs & 300 & 500 \\
Optimizer & AdamW  & AdamW  \\
EMA & 0.9999 & 0.9999 \\
$B$ & 64 & 128 \\
$K$ & 16 & 32  \\
$R$ & 2 & 2 \\
Learning Rate & $1e^{-4}$ & $1e^{-4}$ \\
Weight Decay & 0.1 & 0.1 \\
\hline
\end{tabular}
\caption{Hyperparameter setting for experiments in Tables 1 \& 2 in the main paper.}
\label{tab:hyperparams}
\end{table}

\section{Algorithm for CMM Objective}
The algorithm for calculating the CMM objective ($\mathcal{L}_{CMM}$) is presented below. Note that the model is trained jointly on the combination of flow-matching and $\mathcal{L}_{CMM}$ objectives.

\begin{algorithm}[!h]
\caption{CMM Objective Calculation}
\begin{algorithmic}[1]
\STATE Sample batch of size $K$: $\quad X_0^k \sim \mathcal{N}(0, I), \quad X_1^k \sim D, \quad (d^k, s^k) \sim p(d, s), \quad k = 1, \dots, K$
\STATE $\mathcal{L}_{CMM} \gets 0$
\FOR{$k = 1$ to $K/R$}
    \STATE $X_{\text{few}}^k \gets (1 - s^k) X_0^k + s^k X_1^k$
    \STATE $X_{\text{base}}^k \gets X_{\text{few}}^k$
    \FOR{$r = 1$ to $R$}
        \STATE $\mathcal{S}_1^k \gets \texttt{stopgrad}\big(\mathcal{S}_\theta(X_{\text{base}}^k, s^k, d^k)\big)$
        \STATE $X_{\text{base}}^k \gets X_{\text{base}}^k + \mathcal{S}_1^k \cdot d^k$
        \STATE $\mathcal{S}_2^k \gets \texttt{stopgrad}\big(\mathcal{S}_\theta(X_{\text{base}}^k, s^k+d^k, d^k)\big)$
        \STATE $X_{\text{base}}^k \gets X_{\text{base}}^k + \mathcal{S}_2^k \cdot d^k$ \COMMENT{two half-steps: base surrogate, now at $s^k+2d^k$}
        \STATE $X_{\text{few}}^k \gets X_{\text{few}}^k + \mathcal{S}_\theta(X_{\text{few}}^k, s^k, 2d^k) \cdot 2d^k$ \COMMENT{single full-step: few-step trajectory}
        \STATE $\Delta X \gets X_{\text{few}}^k - X_{\text{base}}^k$ 
        \STATE $\mathcal{L}_{CMM} \gets \mathcal{L}_{CMM} + \big\|\Delta X\|^2$
        \STATE $s^k \gets s^k + 2d^k$
    \ENDFOR
\ENDFOR
\RETURN $\mathcal{L}_{CMM}$
\end{algorithmic}
\label{app:algorithm}
\end{algorithm}

\section{Proof of Lemma 2}

We have,

\begin{align}
 \text{few-step:}\quad &dX^u_t = \big[b(X^u_t, t) + u_\theta(X^u_t, t)\big]\, dt \nonumber \\
 \text{base model:}\quad &dX_t = b(X_t, t)\, dt \nonumber
\end{align}

Let
\[
\Delta X_t = X^u_t - X_t,
\]
which denotes the difference between the states at time $t$ of the few-step and base models starting from the same initial state $x_0\sim p_0$. Using Monge's formulation, the Wasserstein distance between the marginals of the base model and the few-step model is bounded as (with equality if $T(X)=X_t+\Delta X_t$ is the optimal transport map)

\begin{align}
\label{eq:bound}
\mathcal{W}_2^2(p_t^u,p_t) \le \frac{1}{2}\int_X \|\Delta X_t\|^2\,dp_0(x_0) = \frac{1}{2}\mathbb{E}_{x_0}\!\left[\|\Delta X_t\|^2\right]
\end{align}

Our objective is to minimize the cumulative Wasserstein distance between the marginal distributions over time:

\begin{align}
\label{eq:cmm_derive}
\int_{ s}^1 \mathcal{W}_2^2(p_t^u,p_t)\,dt \le \frac{1}{2}\int_{ s}^1 \mathbb{E}_{x_0}\!\left[\|\Delta X_t\|^2\right]dt.
\end{align}

Since $\Delta X_{s}=0$,

\begin{align}
\|\Delta X_t\|^2 = \int_{ s}^{t}\frac{d}{ds}\|\Delta X_s\|^2\,ds.
\end{align}

Substituting into the objective and changing the order of integration,

\begin{align}
\mathcal{L}_{CMM} &= \frac{1}{2}\int_{ s}^1\int_{ s}^{t}\mathbb{E}_{x_0}\!\left[\frac{d}{ds}\|\Delta X_s\|^2\right]ds\,dt \nonumber \\
&= \frac{1}{2}\int_{ s}^1(1-s)\mathbb{E}_{x_0}\!\left[\frac{d}{ds}\|\Delta X_s\|^2\right]ds
\end{align}

Using the chain rule,

\begin{align}
\frac{d}{dt}\|\Delta X_t\|^2 = 2\,\Delta X_t^\top\frac{d\Delta X_t}{dt}
\end{align}

The difference dynamics are

\begin{align}
\Delta X_t &= X_t^u-X_t, \nonumber \\
\therefore\quad \frac{d\Delta X_t}{dt} &= \frac{dX_t^u}{dt}-\frac{dX_t}{dt} \nonumber \\
&= b(X_t^u,t)+u_\theta(X_t^u,t)-b(X_t,t) \nonumber \\
&= \Delta b_t+u_\theta(X_t^u,t),
\end{align}

where

\[
\Delta b_t = b(X_t^u,t)-b(X_t,t)
\]

Substituting into the objective,

\begin{align}
\mathcal{L}_{CMM} &= \frac{1}{2}\mathbb{E}_{x_0}\Bigg[\int_{ s}^1 2\,\Delta X_t^\top\Big(\Delta b_t+u_\theta(X_t^u,t)\Big)(1-t)\,dt\Bigg] \nonumber \\
&= \mathbb{E}_{x_0}\Bigg[\int_{ s}^1 \Delta X_t^\top u_\theta(X_t^u,t)(1-t)\,dt\Bigg] \nonumber \\
&\quad+\mathbb{E}_{x_0}\Bigg[\int_{ s}^1 \Delta X_t^\top\Delta b_t(1-t)\,dt\Bigg].
\end{align}

\section{Proof of Theorem 3}

Since the two trajectories are initialized from the same state $X_s$,
\begin{align}
\Delta X_s = 0.
\end{align}

Moreover,
\begin{align}
\frac{d\Delta X_t}{dt} = \Delta b_t + u_\theta(X_t^u,t),
\end{align}
where
\begin{align}
\Delta b_t = b(X_t^u,t)-b(X_t,t).
\end{align}

Assume that the velocity fields $b(\cdot,t)$ and $u_\theta(\cdot,t)$ are continuously differentiable and Lipschitz continuous in both space and time. Then, over a sufficiently small interval $[s,s+d]$, the trajectory deviation admits the expansion

\begin{align}
\Delta X_t = u_\theta(X_s,s)(t-s) + R_t,
\end{align}

Define the remainder
\begin{align}
R_t = \Delta X_t - u_\theta(X_s,s)(t-s)
\end{align}
Using the integral representation of $\Delta X_t$,
\begin{align}
\Delta X_t = \int_s^t \left[\Delta b_\tau + u_\theta(X_\tau^u,\tau)\right] d\tau,
\end{align}
we obtain
\begin{align}
R_t = \int_s^t \left[\Delta b_\tau + u_\theta(X_\tau^u,\tau) - u_\theta(X_s,s)\right] d\tau
\end{align}

We first bound the two terms in the integrand. Assume $b(\cdot,t)$ is $L_b$-Lipschitz,
\begin{align}
\|\Delta b_\tau\| = \|b(X_\tau^u,\tau)-b(X_\tau,\tau)\| \le L_b\|\Delta X_\tau\|
\end{align}
Furthermore, from Gr\"onwall's inequality,
\begin{align}
\|\Delta X_\tau\| \le Ue^{L_bd}(\tau-s)=\mathcal O(d),
\end{align}
which implies
\begin{align}
\|\Delta b_\tau\|=\mathcal O(d)
\end{align}

Similarly, using the Lipschitz continuity of $u_\theta$ in both state and time,
\begin{align}
\|u_\theta(X_\tau^u,\tau)-u_\theta(X_s,s)\| \le L_u\|X_\tau^u-X_s\|+L_t(\tau-s)
\end{align}
Since the velocity fields are bounded,
\begin{align}
\|X_\tau^u-X_s\| \le (B+U)(\tau-s)=\mathcal O(d),
\end{align}
and therefore
\begin{align}
\|u_\theta(X_\tau^u,\tau)-u_\theta(X_s,s)\|=\mathcal O(d)
\end{align}
Substituting these bounds into the expression for $R_t$ gives
\begin{align}
\|R_t\| &\le \int_s^t \left(\|\Delta b_\tau\|+\|u_\theta(X_\tau^u,\tau)-u_\theta(X_s,s)\|\right)d\tau \nonumber \\
&\le \int_s^t \mathcal O(d)\,d\tau \nonumber \\
&= \mathcal O(d^2)
\end{align}
Hence,
\begin{align}
\sup_{t\in[s,s+d]}\|R_t\|=\mathcal O(d^2).
\end{align}

Substituting this expansion into the restricted CMM objective gives

\begin{align}
\mathcal L_{CMM}[s,s+d] &= \mathbb E\left[\int_s^{s+d}\|u_\theta(X_s,s)(t-s)+R_t\|^2dt\right] \\
&= \mathbb E\left[\int_s^{s+d}\Big(\|u_\theta(X_s,s)\|^2(t-s)^2 + 2(t-s)u_\theta(X_s,s)^\top R_t + \|R_t\|^2\Big)dt\right]
\end{align}

Since $\|R_t\|=\mathcal O(d^2)$,

\begin{align}
\int_s^{s+d}2(t-s)u_\theta^\top R_t\,dt = \mathcal O(d^4) \quad\text{and} \quad
\int_s^{s+d}\|R_t\|^2dt = \mathcal O(d^5)
\end{align}

Therefore,

\begin{align}
\mathcal L_{CMM}[s,s+d] &= \mathbb E\left[\|u_\theta(X_s,s)\|^2\right]\int_s^{s+d}(t-s)^2dt + \mathcal O(d^4) \\
&= \frac{d^3}{3}\mathbb E\left[\|u_\theta(X_s,s)\|^2\right] + \mathcal O(d^4),
\end{align}

which proves the theorem.

\section{Proof of Corollary 4}

Partition the interval $[s,1]$ into
\begin{align}
N=\frac{1-s}{d}
\end{align}
subintervals of equal width $d$.

Applying Theorem 3 on each interval gives

\begin{align}
\mathcal L_{CMM} &= \sum_{k=0}^{N-1}\mathcal L_{CMM}[t_k,t_k+d] \\
&= \frac{d^3}{3}\sum_{k=0}^{N-1}\mathbb E\left[\|u_\theta(X_{t_k},t_k)\|^2\right] + N\mathcal O(d^4).
\end{align}

Since $N=\mathcal O(d^{-1})$,

\begin{align}
N\mathcal O(d^4)=\mathcal O(d^3).
\end{align}

Furthermore,

\begin{align}
d\sum_{k=0}^{N-1}\mathbb E\left[\|u_\theta(X_{t_k},t_k)\|^2\right] \rightarrow J_{CSL},
\end{align}

where the convergence follows from the Riemann sum approximation of the Shortcut-CSL objective. Therefore,

\begin{align}
\mathcal L_{CMM} = \frac{d^2}{3}J_{CSL} + \mathcal O(d^3)
\end{align}

which proves the corollary.

\section{Proof of Theorem 5}

Let the flow be defined by
\begin{align}
\frac{d x_t}{dt} = v_\theta(x_t,t),
\end{align}
and assume that $v_\theta$ is $L$-Lipschitz in its spatial argument, i.e.,
\begin{align}
\|v_\theta(x,t)-v_\theta(y,t)\|
\le
L\|x-y\|
\end{align}
for all $x,y$ and $t$.
Consider two trajectories $x_t$ and $y_t$ with initial conditions
$x_0$ and $y_0$. Define
\begin{align}
d_t := \|x_t-y_t\|.
\end{align}
Then
\begin{align}
\frac{d}{dt}(x_t-y_t)
=
v_\theta(x_t,t)-v_\theta(y_t,t),
\end{align}
and therefore
\begin{align}
\frac{d}{dt} d_t
\le
\|v_\theta(x_t,t)-v_\theta(y_t,t)\|
\le
L d_t.
\end{align}
By Grönwall's inequality,
\begin{align}
d_t \le e^{Lt} d_0.
\end{align}
Substituting back,
\begin{align}
\|x_t-y_t\|
\le
e^{Lt}\|x_0-y_0\|.
\end{align}
Hence the flow map $\psi_t$, defined by $x_t=\psi_t(x_0)$, satisfies
\begin{align}
\|\psi_t(x)-\psi_t(y)\|
\le
e^{Lt}\|x-y\|.
\end{align}
Therefore the Lipschitz constant of the flow map is bounded by
\begin{align}
\operatorname{Lip}(\psi_t)
\le
e^{Lt}.
\end{align}
In particular, for a flow matching model integrated over
$t\in[0,1]$,
\begin{align}
\operatorname{Lip}(\psi_{0\to1})
\le
e^{L}.
\end{align}

\section{Proof of Corollary 6}

Define the $k$-th Euler update
\begin{align}
T_k(x)
=
x+\frac{1}{n}v_\theta(x,t_k).
\end{align}
Since $v_\theta(\cdot,t_k)$ is $L$-Lipschitz,
\begin{align}
\|T_k(x)-T_k(y)\|
&=
\left\|
x-y+\frac{1}{n}
\bigl(v_\theta(x,t_k)-v_\theta(y,t_k)\bigr)
\right\| \\
&\le
\left(1+\frac{L}{n}\right)\|x-y\|.
\end{align}
Hence
\begin{align}
\operatorname{Lip}(T_k)
\le
1+\frac{L}{n}.
\end{align}
Since $\psi^{(n)}=T_{n-1}\circ\cdots\circ T_0$, the Lipschitz constant
of a composition is bounded by the product of the individual
Lipschitz constants:
\begin{align}
\operatorname{Lip}(\psi^{(n)})
&\le
\prod_{k=0}^{n-1}
\operatorname{Lip}(T_k) \\
&\le
\left(1+\frac{L}{n}\right)^n.
\end{align}
Using the classical inequality
\begin{align}
\left(1+\frac{L}{n}\right)^n \le e^L,
\end{align}
we obtain:
\begin{align}
\operatorname{Lip}(\psi^{(n)})
\le
\left(1+\frac{L}{n}\right)^n
\le
e^L.
\end{align}
Finally, the sequence
$\left(1+\frac{L}{n}\right)^n$
is monotonically increasing in $n$, implying that reducing the number of generation steps decreases the upper bound on the Lipschitz constant.

\end{document}